\documentclass[10pt,twocolumn,letterpaper]{article}

 \usepackage[pagenumbers]{iccv} 

\usepackage{amsthm,amsmath,amssymb}
\usepackage{bm}
\usepackage{graphicx}

\definecolor{iccvblue}{rgb}{0.21,0.49,0.74}
\usepackage[pagebackref,breaklinks,colorlinks,allcolors=iccvblue]{hyperref}
\usepackage{color}

\usepackage{pifont}
\usepackage{xcolor} 
\usepackage{tabularx}
\usepackage{booktabs}

\title{MotionDiff: Training-free Zero-shot Interactive Motion Editing 
\\ via Flow-assisted Multi-view Diffusion}

\author{Yikun Ma\\
Sun Yat-sen University\\
{\tt\small mayk25@mail2.sysu.edu.cn}
\and
Yiqing Li\\
Sun Yat-sen University\\
{\tt\small liyq265@mail2.sysu.edu.cn}
\and
Jiawei Wu\\
Sun Yat-sen University\\
{\tt\small wujw97@mail2.sysu.edu.cn}
\and
Xing Luo \\
Department of Frontier Research, Peng Cheng Laboratory, \\
{\tt\small luox@pcl.ac.cn}
\and
Zhi Jin \footnote{Corresponding author}\\
Sun Yat-sen University, \\
Guangdong Provincial Key Laboratory of Fire Science and Intelligent Emergency Technology\\
{\tt\small jinzh26@mail.sysu.edu.cn}
}

\begin{document}
\maketitle

\begin{abstract}
    Generative models have made remarkable advancements 
    and are capable of producing high-quality content. 
    However, performing controllable editing with generative models remains challenging,  
    due to their inherent uncertainty in outputs. 
    This challenge is praticularly pronounced in motion editing, 
    which involves the processing of spatial information. 
    While some physics-based generative methods have attempted to implement motion editing, 
    they typically operate on single-view images with simple motions, such as translation and dragging. 
    These methods struggle to handle complex rotation and stretching motions and ensure multi-view consistency, 
    often necessitating resource-intensive retraining. 
    To address these challenges, we propose MotionDiff, 
    a training-free zero-shot diffusion method that leverages optical flow
    for complex multi-view motion editing. 
    Specifically, given a static scene, users can interactively select objects of interest to add motion priors. 
    The proposed Point Kinematic Model (PKM) then estimates corresponding multi-view optical flows 
    during the Multi-view Flow Estimation Stage (MFES). 
    Subsequently, these optical flows are utilized to generate multi-view motion results 
    through decoupled motion representation in the Multi-view Motion Diffusion Stage (MMDS). 
    Extensive experiments demonstrate that MotionDiff outperforms other physics-based generative motion editing methods in achieving 
    high-quality multi-view consistent motion results. 
    Notably, MotionDiff does not require retraining, 
    enabling users to conveniently adapt it for various down-stream tasks.
\end{abstract}

\section{Introduction}
Generative models, including Generative Adversarial Networks (GANs) \cite{goodfellow2014generative} 
and Diffusion models \cite{rombach2022high, zhang2023adding}, 
possess the capability to produce high-quality content. 
Nevertheless, generating controllable results and executing precise pixel-level 
editing with these models remains challenge. 

Recently, several image editing studies \cite{mokady2023null, epstein2023diffusion, brooks2023instructpix2pix} 
have employed text prompts to execute static editing tasks, 
such as style transfer and object replacement using generative models. 
However, accurately processing motion information based on text prompts poses difficulties, 
leading to suboptimal performance. 
To address this issue, some methods \cite{li2024generative, li2024dragapart, mou2024diffeditor} 
introduce user-defined physical priors, such as dragging and motion points, 
to guide generative models for motion editing. 
Nonetheless, these approaches are generally limited to simple translational dragging 
and are less effective in handling complex motions, such as scaling and rotation. 
In contrast, Motion Guidance \cite{gengmotion} proposes utilizing optical flow 
to guide a diffusion model for complex motion editing. 
However, inadequate supervision of motion objects results in altered texture details. 
Additionally, the absence of multi-view motion constraints 
leads to inconsistent editing performance across different views. 
More recently, certain methods \cite{wang2024diff, wu2025neural, yenphraphai2024image} 
generate physical priors from single images for 3D or multi-view editing. 
However, these methods often require retraining or 
rely on specific field representations (\textit{e.g.}, NeRF \cite{mildenhall2020nerf}, 3DGS \cite{kerbl3Dgaussians}), 
which complicates the editing process and incurs additional resource consumption.

In summary, current physics-based generative motion editing techniques confront three principal challenges: 
\textbf{1)} Drag-based methods excel in managing simple translational 
motions, but encounter difficulties with more complex rotation and stretching motions. 
\textbf{2)} Most methods emphasize single-view image priors, 
which obstructs the achievement of multi-view consistent motion editing. 
\textbf{3)} Most methods necessitate retraining or specific field representations, 
thereby escalating computational resource requirements and data collection expenses. 

To address these challenges, 
we propose a training-free zero-shot multi-view motion editing method called \textbf{MotionDiff}. 
Specifically, it comprises two inference stages: 
\textbf{1) Multi-view Flows Estimation Stage (MFES).}
Given a static scene, MFES enables users to conveniently and interactively 
select the object and motion mode of interest for editing. 
We then propose the Points Kinematic Model (PKM) to estimate multi-view optical flows  
based on the motion of 3D point clouds within the static scene. 
\textbf{2) Multi-view Motion Diffusion Stage (MMDS).}
This stage takes multi-view images and the estimated optical flows as inputs, 
guiding a diffusion model to perform motion editing. 
Our core insight is to decouple the motion representation into  
a combination of static background, motion objects, and occluded regions. 
Consequently, we design corresponding guidance strategies without incurring additional computational costs. 
Extensive experiments demonstrate that MotionDiff 
can effectively implement complex motion editing, while maintaining multi-view consistency 
(as shown in Figure \ref{fig:teaser}), surpassing other physics-based generative motion editing methods. 
Notably, our method provides a user-friendly motion editing solution without retraining. 

The main contributions can be summarized as follows: 
\begin{itemize}
    \item [1)]
    We propose MotionDiff, a training-free zero-shot flow-assisted diffusion framework 
    that enables diverse complex multi-view motion editing operations,  
    such as translation, scaling, rotation and stretching. 
    \item [2)]
    A user-friendly interactive framework is developed within MFES, 
    complemented by the proposed PKM, to derive multi-view optical flows from static scenes. 
    \item [3)]
    A decoupled inference-only motion representation is designed in MMDS, 
    facilitating convenience for other relevant tasks, 
    like Augmented Reality (AR), Virtual Reality (VR) and Human-Computer Interaction (HCI). 
\end{itemize}

The rest of paper is organized as follows:
Section \ref{rw} briefly reviews the related works. 
Section \ref{method} introduces the details of the proposed MotionDiff. 
Section \ref{Experiments} provides comparisons and ablation studies results. 
Conclusion and limitations are summarized in Section \ref{Limitations}. 

\section{Related Works}\label{rw}

\subsection{Image Editing with Generative Models}

Generative models possess the ability to produce realistic and high-quality content, 
including images \cite{zhang2023adding, rombach2022high}, videos \cite{khachatryan2023text2video, singermake}, 
and 3D models \cite{pooledreamfusion, ijcai2024p130}. 
Prior studies \cite{abdal2019image2stylegan, zhu2020domain, pan2023drag} have employed GANs \cite{goodfellow2014generative} for image editing tasks. 
Notably, DragGAN \cite{pan2023drag} stands out by introducing key-handle mechanisms that enable precise control over object movement within images. 
Nevertheless, due to the limited generative capabilities of GANs, 
GANs-based methods often encounter challenges in terms of generalization, 
as well as the authenticity and resolution of the generated content. 

The remarkable success of diffusion models \cite{ho2020denoising, songdenoising, rombach2022high} has drawn significant attention to 
image editing \cite{avrahami2022blended, brooks2023instructpix2pix, hertzprompt, kawar2023imagic, mengsdedit}. 
These approaches generally employ target descriptions or feature attention mechanisms for specific editing tasks.  
However, achieving precise pixel-level editing with physical attributes remains a formidable challenge, 
primarily due to the inherent uncertainty in outputs and the inadequate representation of spatial motion information. 

\subsection{Physics-based Generative Motion Editing}
Recently, there have been significant surge of interests in physics-based and interactive generative editing 
\cite{li2024generative, sajnani2024geodiffuser, 
mou2024diffeditor, shi2024dragdiffusion, zhang2025physdreamer, alzayer2024magic}. 
These approaches typically generate physical motion priors, 
such as vibrations, drags and geometric transformations. 
Subsequently, they employ retrained diffusion models to process these priors and produce motion results. 
However, these methods are primarily effective for handling unordered or simple dragging motions, 
but struggle with complex motions due to their limited spatial motion representation capabilities. 
To address these limitations, Motion Guidance \cite{gengmotion} introduces optical flow 
as a motion prior to enable more sophisticated motion editing. 
Nonetheless, its guidance strategy does not adequately account for texture details of the motion objects, 
leading to inconsistent appearance and multi-view performance. 
In contrast, other methods \cite{wang2024diff, wu2025neural, yenphraphai2024image, xie2024physgaussian, luo20243d} 
choose to generate priors (such as depth maps, bounding boxes, \textit{etc.}) from single images, 
then utilize these priors to explicitly or implicitly enhance editing more complex 3D assets and multi-view images. 
However, these methods require retraining from paired datasets \cite{wang2024diff, wu2025neural}, 
or rely on specific NeRF \cite{yenphraphai2024image} or 3DGS \cite{xie2024physgaussian, luo20243d} 
representation fields, which increase computational resource requirements and complicates the practical editing process.

\section{Proposed Method}\label{method}

\subsection{Preliminaries}
\textbf{Denoising Diffusion.}
Denoising Diffusion Probabilistic Models (DDPM) \cite{ho2020denoising} introduces the diffusion model for image generation. 
Using the reparameterization trick, the added noise of input \bm{$x_0$} can
directly be expressed as: 
\begin{eqnarray}\label{eq:noise}
    & \bm{x_t} = \sqrt{\bar{\alpha_t}}\bm{x_0} + \sqrt{1-\bar{\alpha_t}}\bm{\epsilon} , 
    \ \ where \ \bm{\epsilon} \sim \mathcal{N}(0, \textit{I}), 
\end{eqnarray} 
where $\alpha_t$ represents the learned schedule, \bm{$\epsilon$} denotes the noise variable. 
The computational bottleneck of DDPM is the number of denoising timesteps $T$, 
hence, a non-Markovian variant Denoising Diffusion Implicit Models (DDIM) \cite{songdenoising} is introduced to reduce the number of $T$.
\begin{eqnarray}\label{ddim_pre}
    \begin{aligned}
        \bm{\overline{x}_{pre}} & = \sqrt{\overline{\alpha}_{pre}} \ \ \frac{\bm{x_t}-\sqrt{1-\overline{\alpha}_t}\bm{\epsilon_{\theta}(x_t)}}{\sqrt{\overline{\alpha}_t}} \\
    & + \ \sqrt{1-\overline{\alpha}_{pre}-\sigma^2_t} \ \epsilon_{\theta}(\bm{x_t}) + \sigma^2_t\bm{\epsilon} \ , 
    \end{aligned}
\end{eqnarray} 
where $\epsilon_{\theta}(\bm{x_t})$ denotes the denoising network, usually U-Net \cite{ronneberger2015u}, and $\sigma$ is the variance hyperparameter.

\textbf{Guidance Diffusion.}
Obtaining a specific generative result utilizing diffusion, retraining is usually required. 
However, its retraining is challenging due to the large number of parameters and the immense data. 
Fortunately, classifier guidance \cite{dhariwal2021diffusion} has explored a guidance inference paradigm without retraining, 
which is defined as follows: 
\begin{eqnarray}\label{gudiance_form}
    & \bm{\widetilde{\epsilon_{\theta}}}(\bm{x_t}; t; \bm{y}) = \bm{\epsilon_{\theta}}(\bm{x_t}; t; \bm{y}) 
    + \sigma_t \nabla_{x_t} \mathcal{L}(\bm{x_t}), 
\end{eqnarray} 
where \bm{$y$} denotes the optional conditioning signal, and $\mathcal{L}(\bm{x_t})$ represents the optimization function.

\subsection{Method Overview}

In general, MotionDiff achieves motion editing through two inference stages. 
Firstly, given a static scene, 
users can interactively select the object to be edited and apply motion priors, 
and then the proposed PKM estimates multi-view optical flows within MFES. 
Subsequently, MMDS utilizes these optical flows to 
guide the Stable Diffusion (SD) \cite{rombach2022high} for motion editing 
and decouples the motion representation to obtain multi-view consistent editing results. 
The details of MFES and MMDS are described in Section \ref{MFE_sec} and \ref{MMD_sec}. 

\begin{table}[t]
    \centering 
    \footnotesize
	\caption{Nomenclature}\vspace{-0.3cm}
	\begin{tabularx}{0.46\textwidth}{p{0.48cm}p{2.8cm}p{0.34cm}p{2.95cm}}
		\toprule	
		\bm{$K$} & Intrinsic camera matrix &  \bm{$T$} & Camera translation matrix \\ 
        \bm{$R$} & Camera rotation matrix  & \bm{$Rot$} & Spatial rotation matrix  \\
		\bm{\textcolor{blue}{$P_o$}}  & Original 3D points & \bm{\textcolor{red}{$P_m$}}  & 3D points after motion\\ 
        \bm{$P_{so}$}   & 3D \textbf{s}parse \textbf{o}riginal points & \bm{$P_{sm}$}  & 3D \textbf{s}parse \textbf{m}otion points \\
        \bm{$P_{ox,y}$} & $x, y$ components of \textcolor{blue}{$P_o$} &  \bm{$f_s$}   & Single-view optical flow\\ 
        \bm{$f_{sx,y}$} & $x, y$ components of \bm{$f_s$} & \bm{$f_m$}      & Multi-view optical flows\\
        $foc$      & Camera focal length & $pp$       & Camere principal point \\
        $c_{x,y}$ & 2D pixel coordinates & $d$        & 2D depth  \\
		\bottomrule
	\end{tabularx}
    \vspace{-0.3cm}
\end{table}

\begin{figure}[t]
    \centering
    \includegraphics[width=0.995\linewidth]{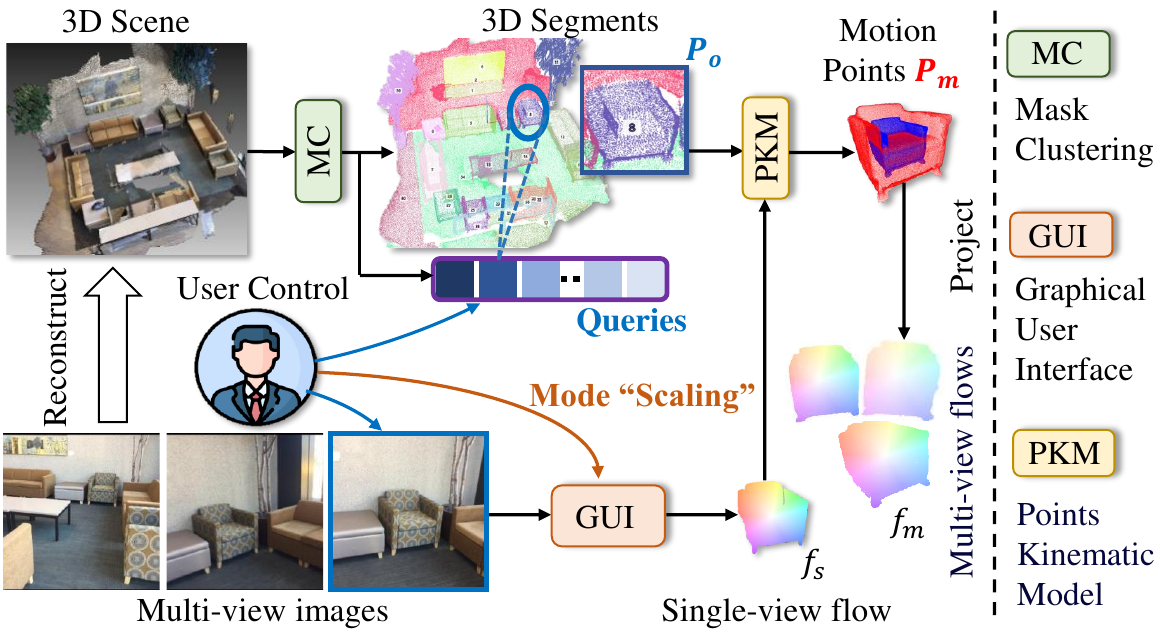}
    \vspace{-0.5cm}
    \caption{The framework of MFES. 
    Given a static scene, 
    users can interactively select a specific-view image from provided multi-view images 
    and the 3D object of interest through queries (\textit{e.g.}, ID 8 represents a ``sofa"). 
    Then the multi-view optical flows are accurately obtained by PKM with user-defined motion mode.} 
    \label{fig:mfe}
    \vspace{-0.5cm}
\end{figure} 

\subsection{Multi-view Flows Estimation Stage} \label{MFE_sec}
Optical flow possesses the ability to represent the pixel-level motion. 
However, directly obtaining the optical flow from a static scene 
with no motion properties is highly challenging. 
Therefore, we propose MFES, allowing users to interactively obtain multi-view optical flows, 
as shown in Figure \ref{fig:mfe}. 
Specifically, given a static scene, 
including multi-view images and reconstructed 3D point clouds (hereinafter referred to as ``3D points"), 
we firstly utilize Mask Clustering \cite{yan2024maskclustering} 
to segment the 3D points and select the \bm{\textcolor{blue}{$P_o$}} of interest through user-interactive queries.  

After acquiring \bm{\textcolor{blue}{$P_o$}}, 
\textbf{our core insight is to estimate the 3D points \bm{\textcolor{red}{$P_m$}} after motion}. 
In this way, we can obtain the corresponding multi-view flows \bm{$f_m$} 
by projecting \bm{\textcolor{red}{$P_m$}} and \bm{\textcolor{blue}{$P_o$}} to the 2D space: 
\begin{eqnarray}\label{touying}
     &\bm{f_m}  =  foc \cdot \bm{[R|T]}^{-1} \odot (\frac{\bm{\textcolor{red}{P_m}}}{d} - \frac{\bm{\textcolor{blue}{P_o}}}{d}) . 
\end{eqnarray}

\bm{\textcolor{blue}{$P_o$}} can be easily obtained utilizing the existing data,  
while it is challenging to obtain \bm{\textcolor{red}{$P_m$}}. 
Therefore, we propose the \textbf{Points Kinematic Model (PKM)} 
\footnote{For more formulas details, please refer to the supplementary materials.}
to estimate \bm{\textcolor{red}{$P_m$}} for different motion modes, 
including translation, scaling, rotation, and stretching, as shown in Figure \ref{fig:pkm}. 
Specifically, we project the selected image to obtain the 3D sparse original points \bm{$P_{so}$}. 
Meanwhile, we design a GUI to interactively generate a single-view optical flow \bm{$f_s$}, 
and the 3D sparse motion points \bm{$P_{sm}$} are represented by \bm{$f_s$}: 
\begin{eqnarray}\label{psm}
    \bm{P_{sm}}\! = \! d \bm{K [R|T]}  [\frac{c_x+\bm{f_{sx}}-pp}{foc},\frac{c_y+\bm{f_{sy}}-pp}{foc},1].\! \! 
\end{eqnarray}

Based on the above definitions, 
we introduce the details of PKM for different motion modes (refer to Figure \ref{fig:pkm}): 

\begin{figure}[tbp]
    \centering
    \includegraphics[width=0.99\linewidth]{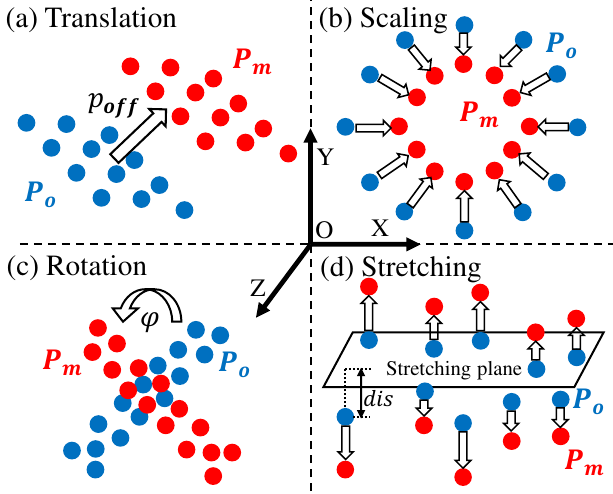}
    \vspace{-0.3cm}
    \caption{The diagrams of different motion modes in PKM.} 
    \label{fig:pkm}
    \vspace{-0.5cm}
\end{figure} 
\textbf{1) Translation.}
The translational motion of 3D points is spatially represented 
as a 3D offset \bm{$p_{off}$} between \bm{\textcolor{red}{$P_m$}} and \bm{\textcolor{blue}{$P_o$}}. 
We contemplate utilizing \bm{$P_{sm}$} and \bm{$P_{so}$} to estimate this \bm{$p_{off}$} based on Eq. (\ref{pingyi}): 
\vspace{-0.3cm}
\begin{eqnarray}\label{pingyi}
    \bm{\textcolor{red}{P_m}}  = \bm{\textcolor{blue}{P_o}} + \bm{p_{off}}, \ \ 
    \bm{p_{off}}  = \frac{\sum_{i=0}^{N(\bm{P_{so}})} (\bm{P_{sm}} - \bm{P_{so}})}{N(\bm{P_{so}})} ,  
\end{eqnarray} 
where $N(\bm{P_{so}})$ denotes the number of \bm{$P_{so}$}.

\textbf{2) Scaling.}
We assume that there exists a scaling factor $s_f$ 
between \bm{\textcolor{blue}{$P_o$}} and \bm{\textcolor{red}{$P_m$}}.
Therefore, we calculate the $s_f$ based on the single-view optical flow \bm{$f_s$}. 

For \textbf{shrinkage}, the $s_f$ can be expressed as: 
\begin{eqnarray}
    s_f = \frac{\sum_{i=0}^{N} ||\bm{f_s}||} { N \cdot Max(||\bm{f_s}||)}, \ \ 
    ||\bm{f_s}|| = \sqrt{{\bm{f_{sx}}}^2 + {\bm{f_{sy}}}^2} , 
\end{eqnarray} 
where $Max(\cdot)$ denotes the maximum value, 
while $N$ represents the number of non-zero optical flow values. 

For \textbf{enlargement}, we obtain the $s_f$ by calculating the ratio of the occluded region \bm{$o_r$}
to the \bm{$f_s$}. Therefore, we need to create the \bm{$o_r$}, and the $s_f$ can be calculated: 
\begin{eqnarray}
    s_f  = \frac{\bm{o_r}}{r(\bm{f_s})}, \ \ \bm{o_r} =  L(c_x, c_y, \bm{f_s}) \cup r(\bm{f_s}) , 
\end{eqnarray} 
where $L(\cdot)$ represents the linear sampling function, 
$r(\cdot)$ represents the region of non-zero optical flow. 

Thus, the \bm{\textcolor{red}{$P_m$}} of above scaling motion can be defined as:
\begin{eqnarray}
    \bm{\textcolor{red}{P_m}}  &=& s_f \cdot \bm{\textcolor{blue}{P_o}} . 
\end{eqnarray} 

\textbf{3) Rotation.} 
For rotation, all 3D points have the same rotation angle $\varphi$. 
Fortunately, we can conveniently obtain $\varphi$ based on the designed GUI. 
Therefore, the spatial rotation matrix \bm{$Rot$} can be defined as: 
\vspace{-0.2cm}
\begin{equation}
    \bm{Rot} =
	\begin{bmatrix}
	 cos  \varphi & -sin  \varphi & 0 \\
     sin \varphi & cos  \varphi & 0 \\
	0 & 0 & 1
	 \end{bmatrix}, 
\end{equation}
thus, the \bm{\textcolor{red}{$P_m$}} of rotation can be defined as:
\begin{eqnarray}
    \bm{\textcolor{red}{P_m}}  &=& \bm{Rot} \odot (\bm{\textcolor{blue}{P_o}} - \bm{p_c}) + \bm{p_c} \ ,  \vspace{-0.1cm}
\end{eqnarray}
where \bm{$p_c$} represents the centroid of \bm{\textcolor{blue}{$P_o$}}.

\textbf{4) Stretching.}
For stretching, we assume \bm{\textcolor{blue}{$P_o$}} are segmented by 
a 2D stretching plane and are stretched to varying degrees along this plane. 
Since the \bm{$f_s$} is obtained in the $XY$ plane, 
the stretching plane is perpendicular to $XY$ plane and parallel to $Z$-axis: 
\begin{eqnarray}
    Ax + By + D = 0 .
\end{eqnarray} 

Therefore, we can find two 3D points $\bm{P_1} = (x_1, y_1, z_1), \bm{P_2} = (x_2, y_2, z_2)$ located on this plane, 
thus calculating the unique 2D stretching plane: 
\begin{align}\left\{\begin{aligned}
    A & = y_2 - y_1 , \ \ B = x_1 - x_2 ,  \\ 
    D & = (x_2 - x_1)y_1 - (y_2 - y_1)x_1 \  \  ,   
\end{aligned}\right.\end{align}
therefore, the distance of \bm{\textcolor{blue}{$P_o$}} to this plane \bm{$dis$} and the stretching factor \bm{$t_f$} are expressed as: 
\begin{eqnarray}
    \bm{dis} = \frac{A\bm{P_{ox}}+B\bm{P_{oy}}+D}{\sqrt{A^2 + B^2}}, \ \ \bm{t_f} = \frac{\bm{dis}}{|Max(\bm{dis})|}. 
\end{eqnarray} 

Therefore, the \bm{\textcolor{red}{$P_m$}} of stretching motion are defined as: 
\begin{eqnarray}
    \bm{\textcolor{red}{P_m}} = \bm{\textcolor{blue}{P_o}} + \bm{t_f} \odot Max(\bm{P_{sm}} - \bm{P_{so}}) .
\end{eqnarray} 

In summary, by designing the PKM, we can estimate the 3D motion points \bm{\textcolor{red}{$P_m$}},   
and subsequently obtain multi-view optical flows \bm{$f_m$} according to \bm{\textcolor{red}{$P_m$}} and \bm{\textcolor{blue}{$P_o$}} in Eq. (\ref{touying}). 

\begin{figure*}[htbp]
    \centering
    \includegraphics[width=0.998\linewidth]{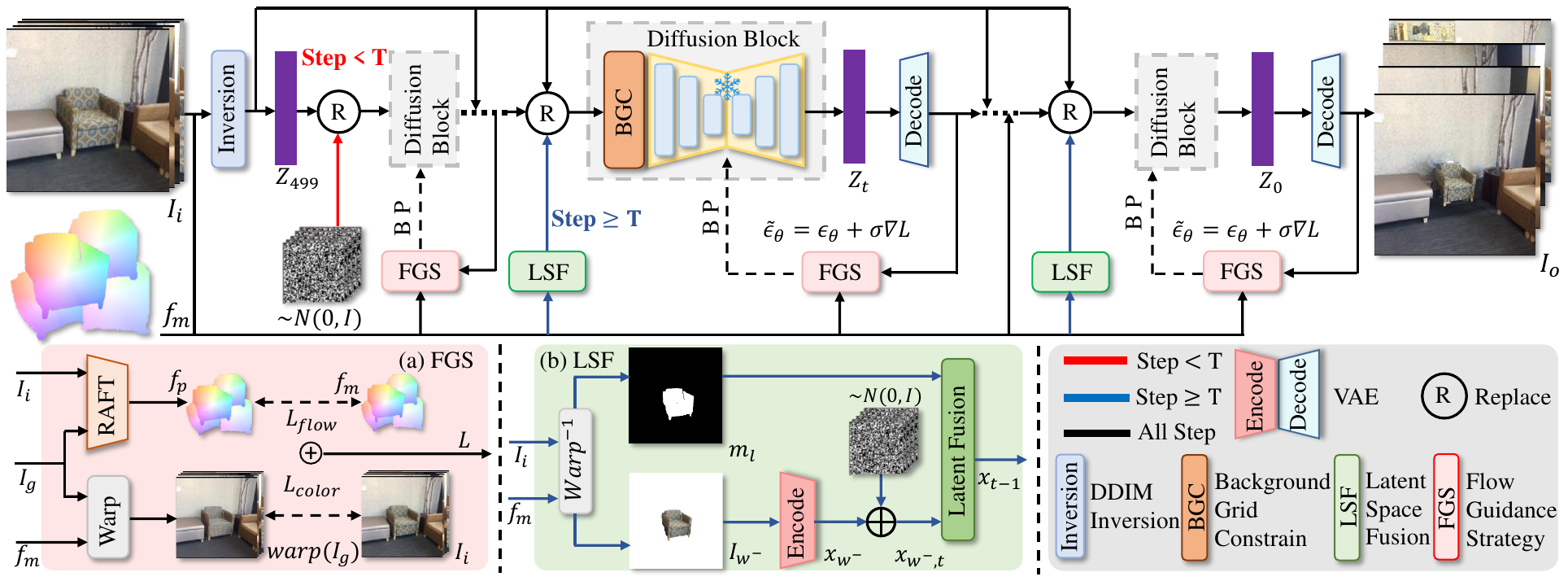}
    \vspace{-0.6cm}
    \caption{
    The pipeline of MMDS. Given multi-view images and their optical flows, 
    we perform DDIM inversion in all diffusion steps to preserve the static background from being tampered with. 
    In the first \textcolor{red}{$T$} steps, only FGS is used as the guidance to ensure a reasonable motion trend. 
    From \textcolor{blue}{$T+1$} step, both FGS and LSF are utilized to obtain 
    corresponding motion with fidelity of texture details. 
    Meanwhile, 
    BGC is embedded into the diffusion model in all steps to maintain multi-view consistency. 
    }
    \label{fig:mmds}
    \vspace{-0.6cm}
\end{figure*} 

\begin{figure}[htbp]
    \centering
    \includegraphics[width=0.99\linewidth]{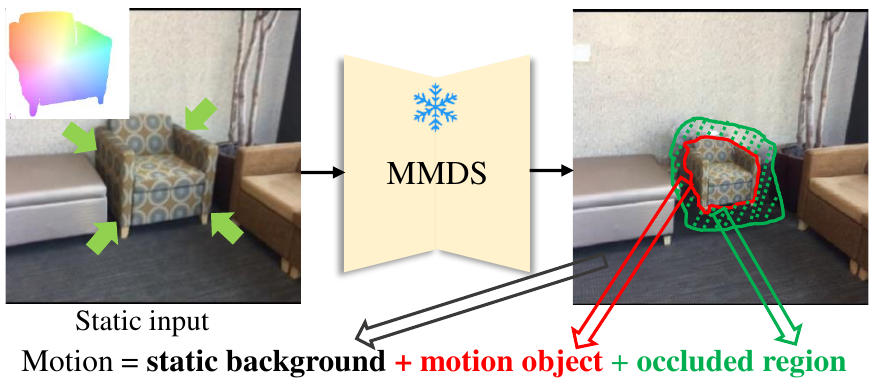}
    \vspace{-0.2cm}
    \caption{The motion decoupled representation.}
    \label{fig:jieou}
    \vspace{-0.4cm}
\end{figure} 

\subsection{Multi-view Motion Diffusion Stage}\label{MMD_sec}

Controlling diffusion models for motion editing often requires retraining, 
which is demanding on computational resources cost and data. 
Therefore, in MMDS (Figure \ref{fig:mmds}), we propose a training-free zero-shot diffusion model, 
leveraging the multi-view optical flows obtained from MFES to 
guide Stable Diffusion (SD) \cite{rombach2022high} in completing motion editing task. 
\textbf{Our core insight is to decouple the motion process into  
a combination of static background, motion objects, and occluded region, as shown in Figure \ref{fig:jieou}.}

Specifically, in all diffusion steps, we utilize DDIM Inversion \cite{songdenoising} to obtain the non-moving region latents 
to prevent the \textbf{static background} structure from being tampered with. 
Meanwhile, we design the Flow Guidance Strategy (FGS) in all steps to guide SD for motion editing. 
From the $T$ step, we introduce the processed Latent Space Fusion (LSF) strategy to maintain the texture details of the generated \textbf{motion objects}. 
To guarantee multi-view consistency of the generated \textbf{occluded region}, 
we introduce the Background Grid Constrain (BGC). 
Finally, we utilize VAE \cite{kingma2013auto} to decode the predicted noise latents and 
obtain multi-view motion results. 
In the following, we describe the details of decoupled motion representation strategies in MMDS: 

\textbf{Flow Guidance Strategy.} 
In FGS, we aim to utilize multi-view optical flows to 
guide SD in zero-shot motion editing. 
Firstly, we employ Recurrent All-Pairs Field Transforms (RAFT) \cite{teed2020raft} 
to predict the optical flows \bm{$f_p$} from the output images \bm{$I_o$} and input images \bm{$I_i$}, 
and then calculate the flow loss $\mathcal{L}_{flow}$. 
Simultaneously, we apply the warp transformation to \bm{$I_o$} and 
compute the color loss $\mathcal{L}_{color}$: 
\begin{eqnarray}\label{eq:loss}
    & \mathcal{L}_{flow} = ||\bm{f_m} - \bm{f_p}||_1 , \ \bm{f_p} = RAFT(\bm{I_i}, \bm{I_o}) , \\
    & \mathcal{L}_{color} = ||\bm{I_i} - warp(\bm{I_o})||_1 .
\end{eqnarray} 

Subsequently, 
these two losses are used to update gradients in Eq. (\ref{gudiance_form}) 
and guide the diffusion for motion editing.

\textbf{Latent Space Fusion.} 
However, we found directly applying FGS leads to changes in the texture details of \textbf{motion objects}, 
as shown in Figure \ref{fig:single_aba}(d). 
Thus, we propose LSF to achieve high-quality texture fidelity without increasing the computational cost. 
Specifically, we first perform inverse warp transformation on \bm{$I_i$} to obtain the target object \bm{$I_{w^-}$} 
and its latent space mask \bm{$m_l$}, 
then encode the \bm{$I_{w^-}$} using a VAE encoder: 
\begin{eqnarray}
    & \bm{I_{w^{-}}} = warp^{-1}(\bm{I_i}, \bm{f_m}) , \ \  
    \bm{x_{w^{-}}} = E(\bm{I_{w^{-}}}) , 
\end{eqnarray} 
where $E$ denotes the VAE encoder.

Then, in accordance with the forward process of DDIM, 
we add noise to the encoded image \bm{$x_{w^{-}}$} based on Eq. (\ref{eq:noise}):
\begin{eqnarray}
    & \bm{x_{w^{-},t}} = \sqrt{a_t} \bm{x_{w^{-}}} + \sqrt{1-a_t} \bm{\epsilon} , 
     \bm{\epsilon} \sim \mathcal{N}(0, \textit{I}), 
\end{eqnarray} 
where $\sqrt{a_t}$ denotes the hyperparameter related to the process of adding noise.
Finally, we perform latent fusion of the forward noise \bm{$\overline{x_{t-1}}$} predicted by the Eq. (\ref{ddim_pre}):
\begin{eqnarray}
    & \bm{x_{t-1}} = \bm{m_l} \odot \bm{x_{w^{-},t}} + (1-\bm{m_l}) \odot \bm{\overline{x_{t-1}}} . 
\end{eqnarray} 

Through LSF, we can maintain the texture details of the motion objects, 
shown in Figure \ref{fig:single_aba}(f).

\begin{figure}[t]
    \centering
    \includegraphics[width=0.99\linewidth]{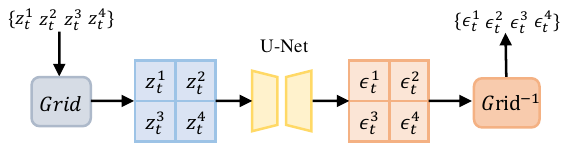}
    \vspace{-0.3cm}
    \caption{The framework of BGC, which can enforce 
    multi-view consistency without introducing additional computational costs.}
    \label{fig:bgc}
    \vspace{-0.5cm}
\end{figure}

\textbf{Background Grid Constrain.}
Benefiting from FGS and LSF strategies, 
we can achieve motion editing with consistent texture details of motion objects. 
However, the lack of effective multi-view consistency constraints 
results in uncontrollable content generation within the \textbf{occluded region}, 
as shown in Figure \ref{fig:aba_bgc}(b). 
Previous works \cite{weber2024nerfiller, bar2022visual} found that using grid constraints 
during the generation process can benefit the multi-view consistency. 
Therefore, we aim to introduce a relevant constraint that 
preserves the consistency of occluded region across different views. 
Hence, we propose the Background Grid Constrain (BGC), 
as shown in Figure \ref{fig:bgc}, which is defined as follows:
\begin{eqnarray}
    & \widetilde{\epsilon_{\theta}}(\bm{x_t}; \ t; \ \bm{y}) = G^{-1}(\epsilon_{\theta}(G(\bm{x_t}); \ t; \ \bm{y})), 
\end{eqnarray} 
where $G$ denotes the grid transformation. 

Specifically, a latent tensor \bm{$\mathcal{Z}$} with the shape of $(4, 4, 64, 64)$ 
undergoes a transformation through $G$, its size changes to $(1, 4, 128, 128)$. 
Then, U-Net \cite{ronneberger2015u} is utilized to predict the noise latent \bm{$\epsilon$}, 
and we perform inverse grid transformation $G^{-1}$ to obtain the latent with shape $(4, 4, 64, 64)$. 
Through this grid-based constrain, BGC can leverage the multi-view information to complement each other,  
thus improving multi-view consistency without increasing the computation cost. 
In summary, through the designed MFES and MMDS, 
our MotionDiff can achieve complex multi-view consistent motion editing for a static scene. 

\section{Experiments}\label{Experiments}
\subsection{Implementations Details}
We implement our method on four 48G memory NVIDIA RTX A6000 GPUs. 
Our experiments are conducted on the widely-recognized ScanNet200 \cite{dai2017scannet, rozenberszki2022language}, 
which comprises of 312 scenes with multi-view images and 3D points.  
For segmenting 3D points and estimating optical flows, 
we utilize the frozen Mask Clustering \cite{yan2024maskclustering} and RAFT \cite{teed2020raft}. 
The SD-1.5 \cite{rombach2022high} model is served as our denoising framework. 
All input and output images are processed at 512 $\times$ 512 resolution. 
DDIM sampling is adopted with 500 steps and the classifier-free guidance scale is set to 7.5. 
To ensure a fair experimental setup, we follow Motion Guidance \cite{gengmotion} protocol, 
incorporating five distinct motion modes. 
Subsequently, we perform both quantitative and qualitative comparisons against benchmark methods, including 
DiffEditor \cite{mou2024diffeditor}, Motion Guidance \cite{gengmotion}, and MagicFixup \cite{alzayer2024magic}.

\subsection{Quantitative Comparisons}
Previous employed image editing metrics, 
such as FID \cite{Seitzer2020FID} and CLIP Similarity \cite{gal2022stylegan, radford2021learning}, 
are inadequate for motion editing task due to the significant 
spatial geometric transformations and the emergence of diverse novel regions. 
Additionally, the absence of paired input-motion datasets 
renders traditional subjective metrics like PSNR and SSIM \cite{wang2004image} unsuitable. 
Consequently, we introduce three multi-view motion editing evaluation metrics: 

\textbf{Motion Position Accuracy (MPA)}, 
calculated by RAFT to represent the motion precision between input and output: 
\begin{eqnarray}
    & MPA = \lambda_{mpa} \ \mathcal{L}_{flow}, 
\end{eqnarray}

\textbf{Appearance Texture Fidelity (ATF)}, 
denoting the extent of texture appearance change in the motion objects: 
\begin{eqnarray}
    & ATF = \lambda_{atf} \ ||\bm{m} \odot \bm{I_{w^{-}}}, \ \bm{m} \odot \bm{I_o}||_1, 
\end{eqnarray}
where \bm{$m$} represents the non-zero regions masks of \bm{$f_m$}.

\textbf{Multi-View Consistency (MVC)},
we adopt Overlapping PSNR proposed by MVDiffusion \cite{NEURIPS2023_a0da690a}:
\begin{eqnarray}
    \textit{MVC} = P(\bm{I_1}, \textit{pt}(\bm{I_2}, \bm{K} \odot \bm{R_1}^{-1} \odot \bm{R_2} \odot \bm{K}^{-1})), 
\end{eqnarray}
where $P$ represents the standard PSNR, \bm{$I_{1,2}$} denote two similar images, 
$pt$ signifies the perspective transformation.

\begin{table}[t]   
    \vspace{-.1cm}     
    \begin{center}
        \footnotesize
        \setlength\tabcolsep{1.6pt}
        \belowrulesep=0pt
        \aboverulesep=0pt
        \begin{tabular}{c|c|c|c|c}
        \toprule
        -     & Translation             & Scaling                & Rotation                 & Stretching        \\
        \midrule
        DE    & 14.7/10.5/33.51         & 15.7/19.4/33.34        & 18.1/14.3/27.45          & 7.4/16.3/36.31    \\
        MG    & 59.5/25.1/29.31         & 53.0/20.1/28.33        & 39.3/14.6/22.84          & 35.7/30.8/35.53   \\
        MF    & 26.5/9.2/33.24          & 10.0/15.6/34.29        & 17.0/14.1/29.78          & 6.2/7.0/37.78     \\
        Ours  & \textbf{10.8/4.6/34.38} & \textbf{6.7/6.2/34.76} & \textbf{10.6/11.1/30.02} & \textbf{3.1/3.4/38.62}  \\
        \bottomrule
        \end{tabular}
        \end{center}
        \vspace{-.35cm}
        \caption{Quantitative comparisons with other physics-based generative motion editing methods. 
        Each column represents one specific motion on 
        \textbf{MPA $\downarrow$ / ATP $\downarrow$  / MVC $\uparrow$ } metrics. 
        The \textbf{DE, MG, MF} represent DiffEditor, Motion Guidance, and MagicFixup, respectively. 
        MotionDiff achieves the SOTA quantitative performance.}
        \label{tab:com}
        \vspace{-.4cm}
\end{table}

In Table \ref{tab:com}, we present quantitative comparisons of each motion modes. 
Other methods exhibit unsatisfactory results in terms of MPA and ATF metrics, 
primarily due to inadequate motion accuracy and compromised appearance fidelity.
Additionally, these methods struggle to maintain multi-view consistency, 
resulting in lower MVC values. 
In contrast, MotionDiff demonstrates SOTA performance across all metrics, 
attributed to its capability in achieving precise texture-consistent multi-view motion editing. 
\begin{figure*}[htbp]
    \centering
    \includegraphics[width=0.98\linewidth]{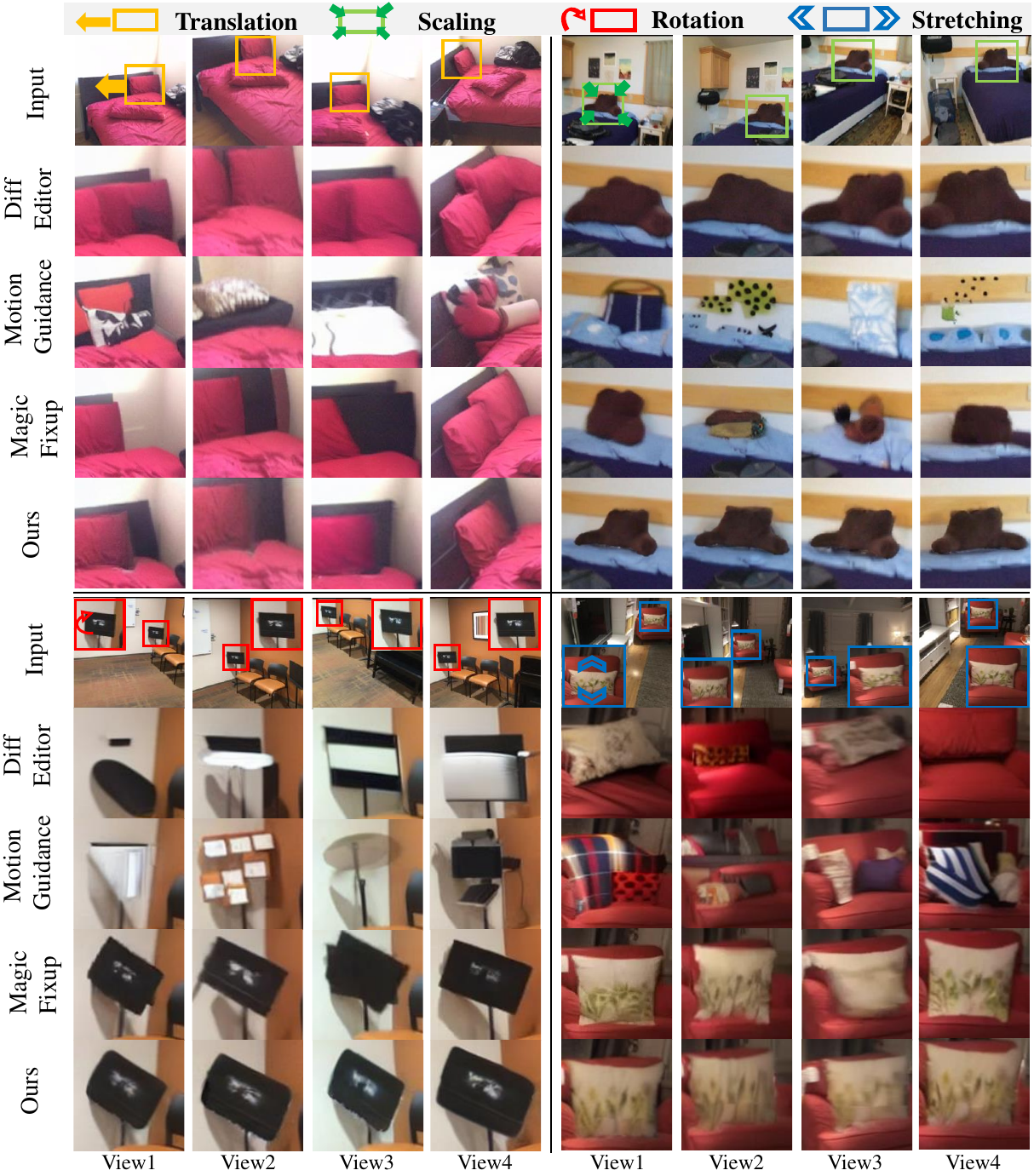}
    \vspace{-0.3cm}
    \caption{Visual comparisons between MotionDiff and other physics-based motion editing methods 
    on translation, scaling, rotation, and stretching tasks.
    Our method achieves better motion editing performance while maintaining multi-view consistency. 
    } 
    \label{multi_com}
    \vspace{-0.5cm}
\end{figure*} 
\begin{figure*}[htbp]
    \centering
    \includegraphics[width=0.98\linewidth]{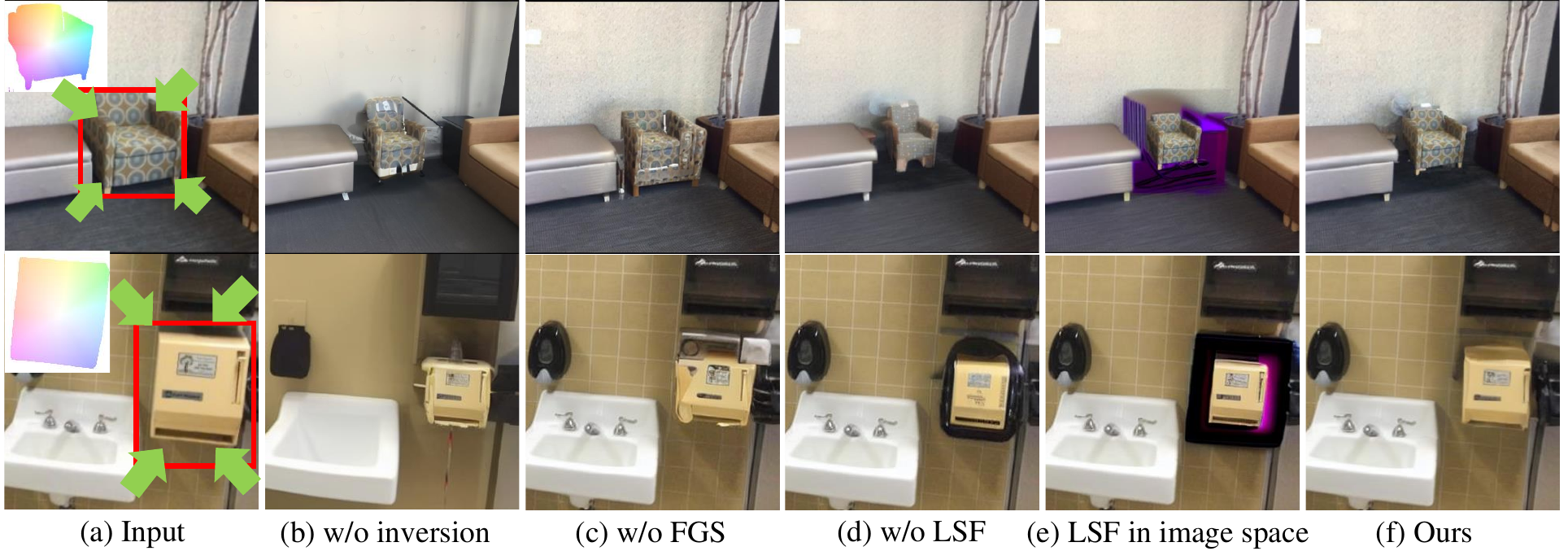}
    \vspace{-0.25cm}
    \caption{Visual comparisons of our ablation studies of MotionDiff.}
    \label{fig:single_aba}
    \vspace{-0.65cm}
\end{figure*} 

\subsection{Qualitative Comparisons}
To comprehensively evaluate the performance of MotionDiff, 
we provide detailed qualitative comparisons against other benchmark methods.  

We input multi-view images, and fairly apply corresponding motions to all methods. 
As illustrated in Figure \ref{multi_com}, we present visual comparisons for 
translation, scaling, rotation, and stretching, respectively. 
Each row showcases the editing outcomes of \textcolor{red}{different methods}, 
while each column represents the \textcolor{blue}{different views from a single scene}. 
To better display the motion results, we zoom in the \textbf{box regions}. 
It demonstrates that 
DiffEditor \cite{mou2024diffeditor} can manage simple motions, 
such as translation and scaling to some extent, 
but is limited in handling more complex motions 
(\textit{e.g.}, the appearance of the blackboard in rotation is unsatisfactory). 
Additionally, Motion Guidance \cite{gengmotion} struggles to maintain the appearance of motion objects, 
due to lacking an effective texture supervision strategy, 
for instance, the texture details of the red pillow are lost during translation. 
Moreover, MagicFixup \cite{alzayer2024magic} achieves satisfactory results in single-view motion editing, 
but its multi-view motion editing performance is limited due to the inadequate consistency constraints. 
In contrast, MotionDiff excels in executing diverse motions,  
while effectively preserving multi-view consistency. 

\subsection{Ablation Studies}\label{sec:aba}
\begin{figure}[t]
    \centering
    \includegraphics[width=0.97\linewidth]{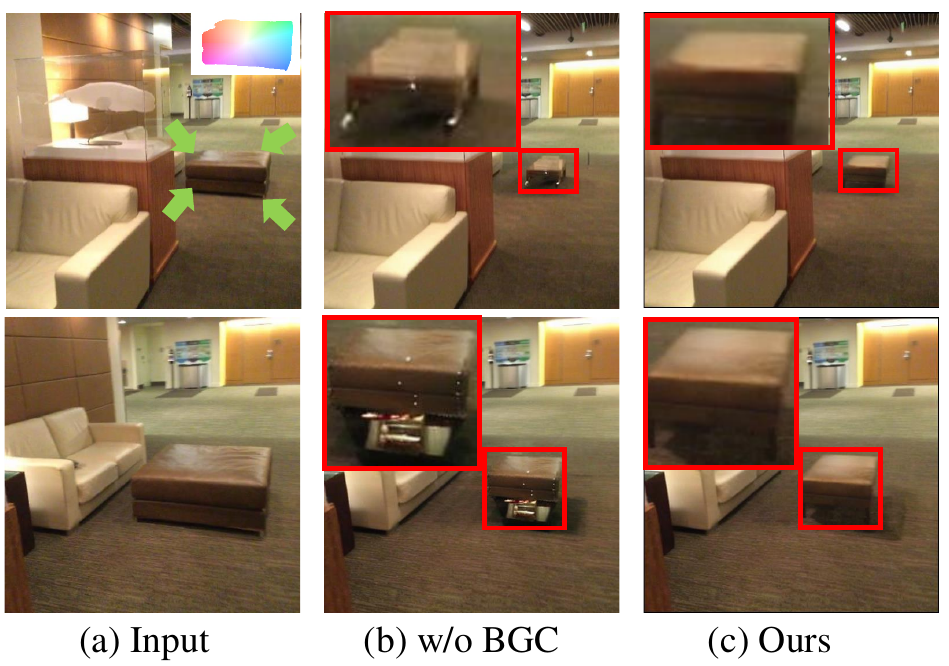}
    \vspace{-0.2cm}
    \caption{Visual ablation study of BGC. 
    It demonstrates that BGC is advantageous in ensuring consistency across multi-view images.}
    \label{fig:aba_bgc}
    \vspace{-0.5cm}
\end{figure}

To validate the efficacy of the proposed strategies, 
we design corresponding ablation studies. 
The visualization and metric outcomes are presented in Figure \ref{fig:single_aba}, \ref{fig:aba_bgc} and Table \ref{tab:aba}. 

\textbf{W/o inversion.} In Figure \ref{fig:single_aba}(b), 
the absence of DDIM inversion substitution
makes it difficult to maintain a \textbf{static background}, 
generating uncontrollable content. 

\textbf{W/o FGS.} As depicted in Figure \ref{fig:single_aba}(c), 
the completion of physical motion becomes challenging without FGS, 
as there is no motion prior provided to the diffusion model. 

\textbf{W/o LSF.} As illustrated in Figure \ref{fig:single_aba}(d), 
the absence of the LSF strategy leads to 
inadequate supervision over \textbf{motion objects}, 
thereby failing to guarantee the texture details. 

\textbf{LSF in image space.} 
To evaluate the effectiveness of the latent fusion strategy, 
we transform it into the image space. 
However, as shown in Figure \ref{fig:single_aba}(e), 
due to the substantial distribution difference between these two spaces, 
the quality of the generated content hardly be guaranteed. 

\textbf{W/o BGC.} As indicated in Figure \ref{fig:aba_bgc}(b), 
the absence of multi-view constraints from the background grid 
results in inconsistent generation of the \textbf{occluded region}. 
\begin{table}[htbp]   
    \begin{center}
        \footnotesize
        \setlength\tabcolsep{1.2pt}
        \belowrulesep=0pt
        \aboverulesep=0pt
        \begin{tabular}{c|c|c|c|c|c|c}
        \toprule
        -       & w/o inversion  & w/o FGS    & w/o LSF  &LSF in image & w/o BGC  & Ours   \\
        \midrule
        MPA $\downarrow$      & 9.5            & 9.4        & 7.7      & 7.7         & 7.2      & \textbf{6.8}    \\
        ATF $\downarrow$       & 5.6            & 10.2       & \textbf{5.1}      & 7.5         & 5.3      & \textbf{5.1}    \\
        MVC  $\uparrow$      & 28.96          & 32.14      & 31.79    & 30.23       & 30.22    & \textbf{34.92}   \\
        \bottomrule
        \end{tabular}
        \end{center}
        \vspace{-.5cm}
        \caption{Quantitative results of the ablation studies.}
        \label{tab:aba}
        \vspace{-.5cm}
\end{table}  
In summary, extensive ablation studies validate the efficacy of the proposed strategies. 
Through their integration, MotionDiff accomplishes sophisticated 
multi-view consistent motion editing without any retraining. 

\subsection{Extension Experiments} 
\begin{figure}[t]
    \centering
    \includegraphics[width=0.99\linewidth]{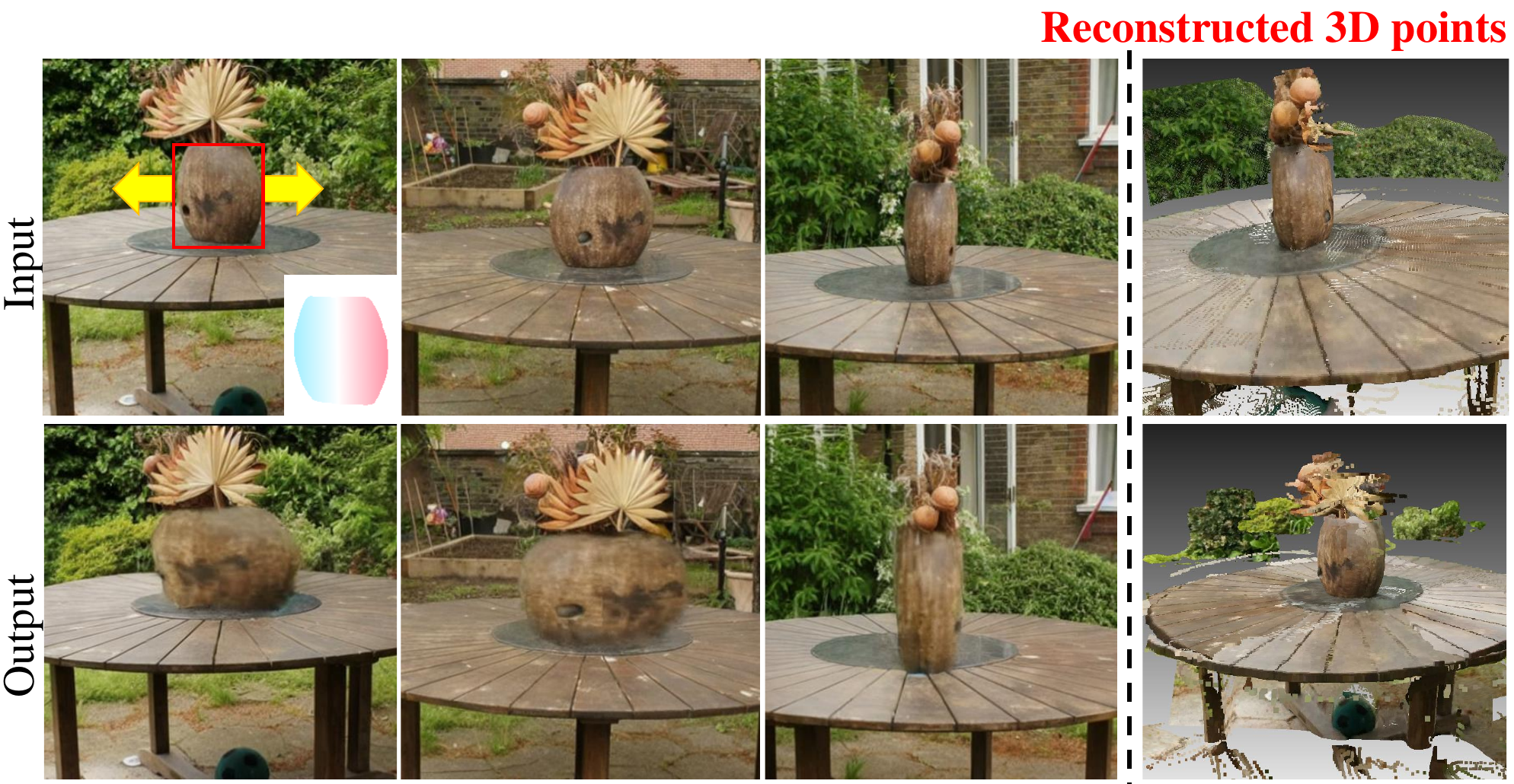}
    \vspace{-0.2cm}
    \caption{Given \textcolor{blue}{sparse-view images}, 
    MotionDiff obtains corresponding editing results with 
    reconstructed 3D points by DUSt3R.}
    \label{fig:tuo}
    \vspace{-0.6cm}
\end{figure}

Given that ScanNet dataset provides 3D points for each scene, we utilize these directly for simplicity. 
\textbf{However, it is important to note that our method can also operate using multi-view images as the only input}, 
with 3D points being reconstructed through established techniques, 
such as DUSt3R \cite{wang2024dust3r} and MV-DUSt3R+ \cite{tang2024mv}. 
To validate this, we select arbitrary \textbf{outdoor sparse-view} 
images from Mip-NeRF 360 \cite{barron2022mip},  
and reconstruct corresponding 3D points using DUSt3R. 
Figure \ref{fig:tuo} demonstrates the motion results, 
further validating the editing practicality of our method. 

\section{Conclusion and Limitations}\label{Limitations}
A training-free zero-shot diffusion framework MotionDiff is proposed, 
facilitating diverse complex multi-view motion editing tasks. 
An convenient interactive framework is designed within MFES to estimate optical flow, 
while the motion process is decoupled in MMDS, 
allowing users to freely adopt it for other down-stream tasks. 
Nonetheless, the proposed PKM somewhat limits the motion modes. 
establishing corresponding models for other motions in PKM is required, 
while we believe it is straightforward to implement. 
In future work, we will focus on integrating more general motion representation for 
generative editing.

{
    \small
    \bibliographystyle{ieeenat_fullname}
    \bibliography{main}
}

\end{document}